\documentclass[letterpaper, 10 pt, conference]{ieeeconf}  % Comment this line out if you need a4paper

\IEEEoverridecommandlockouts                              % This command is only needed if 
\usepackage[left=1.88cm,right=1.88cm,top=1.88cm,bottom=1.88cm]{geometry}
\usepackage{amsmath,amssymb,amsfonts}
\usepackage{cite}
\usepackage{graphicx}
\usepackage{textcomp}
\usepackage{xcolor}
\usepackage{bbm}
\let\labelindent\relax
\usepackage{enumitem}
\usepackage{tabularx}
\usepackage{times}
\usepackage{multicol}

\usepackage{graphicx}
\usepackage{bm}
\usepackage[nolist]{acronym}

\usepackage[linesnumbered,ruled]{algorithm2e}
\usepackage{caption}
\usepackage{mathtools}
\usepackage{array}
\usepackage{subfigure}
\usepackage{diagbox}
\usepackage{etoolbox}\AtBeginEnvironment{algorithmic}{\small}
\usepackage{soul} % Sangjae added, Aug 24, 2021
\usepackage{lipsum}
\usepackage[colorinlistoftodos]{todonotes}

\usepackage{amsthm}

\newtheorem{remark}{Remark}
\newtheorem{lemma}{Lemma}
\newtheorem{example}{Example}
\usepackage{bbm} % for mathbb 1

\usepackage[colorlinks = True, linkcolor = blue, citecolor = blue]{hyperref}
\DeclareMathOperator*{\argmax}{arg\,max}

\title{\LARGE \bf
Active Probing with Multimodal Predictions for Motion Planning
 }

\author{Darshan Gadginmath$^{1,2}$ \quad Farhad Nawaz$^{1}$ \quad Minjun Sung$^{1}$ \quad Faizan M Tariq$^1$  \\ Sangjae Bae$^1$ \quad David Isele$^1$ \quad Fabio Pasqualetti$^2$ \quad Jovin D'sa$^1$ \thanks{Work was conducted during D. Gadginmath’s (\href{mailto:dgadg001@ucr.edu}{\texttt{dgadg001@ucr.edu}}) internship at the $^1$Honda Research Institute, USA, San Jose, CA, 95134.
$^2$Department of Mechanical Engineering, University of California Riverside, Riverside, CA, 92521.}}
\begin{document}

\maketitle
% \thispagestyle{empty}
% \pagestyle{empty}

%%%%%%%%%%%%%%%%%%%%%%%%%%%%%%%%%%%%%%%%%%%%%%%%%%%%%%%%%%%%%%%%%%%%%%%%%%%%%%%%

\begin{abstract}
  Navigation in dynamic environments requires autonomous systems to reason about uncertainties in the behavior of other agents. In this paper, we introduce a unified framework that combines trajectory planning with multimodal predictions and active probing to enhance decision-making under uncertainty. We develop a novel risk metric that seamlessly integrates multimodal prediction uncertainties through mixture models. When these uncertainties follow a Gaussian mixture distribution, we prove that our risk metric admits a closed-form solution, and is always finite, thus ensuring analytical tractability. To reduce prediction ambiguity, we incorporate an active probing mechanism that strategically selects actions to improve its estimates of behavioral parameters of other agents, while simultaneously handling multimodal uncertainties. We extensively evaluate our framework in autonomous navigation scenarios using the MetaDrive simulation environment. Results demonstrate that our active probing approach successfully navigates complex traffic scenarios with uncertain predictions. Additionally, our framework shows robust performance across diverse traffic agent behavior models, indicating its broad applicability to real-world autonomous navigation challenges. Code and videos are available at \url{https://darshangm.github.io/papers/active-probing-multimodal-predictions/}.
\end{abstract}

%%%%%%%%%%%%%%%%%%%%%%%%%%%%%%%%%%%%%%%%%%%%%%%%%%%%%%%%%%%%%%%%%%%%%%%%%%%%%%%%
\section{INTRODUCTION}
% \jd{it would be better to write the entire paper in active voice as much as possible to keep that consistent. I will edit some sections as i find them}.
Autonomous vehicles must navigate complex environments by anticipating and responding to the behavior of other agents \cite{FMT-DI-JSB-SB:2023_slas}. While humans naturally reason about multiple possible future trajectories and adapt their decisions accordingly, replicating this capability in autonomous systems remains challenging. Recent advances in prediction models leverage sophisticated architectures including transformers~\cite{SS-LJ-DD-BS:2024_mtr,NN-etal:2023_wayformer,KZ-XF-LW-ZH:2022}, graph neural networks~\cite{TS-BI-PC-MP:2020}, and denoising diffusion mechanisms~\cite{CL-SH-HL-JC:2024,ZL-etal:2023_diffusion} to generate multimodal trajectory predictions. However, these models face fundamental limitations stemming from their training data. Popular datasets like Waymo Open Motion, NuScenes, and Lyft struggle to capture the full spectrum of human driving styles, while variations in sensor quality during data collection introduce additional uncertainties~\cite{LF-etal:2024_unitraj}. As a result, these models often falter when confronted with out-of-distribution scenarios \cite{JL-JL-SB-DI:2024}, raising critical safety concerns for real-world deployment.

\begin{figure}
\includegraphics[width = 0.50\textwidth]{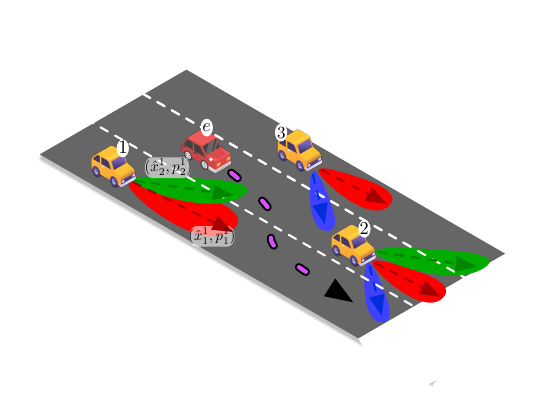} 
\vspace*{-1.2cm}
\caption{Illustrative figure of a merging scenario with uncertain multimodal predictions. The red vehicle represents the ego agent with its intended trajectory given by the pink dashed line. The other agents are represented by the yellow cars. The predictor provides multimodal predictions $\hat{x}^i_k$ of the trajectory of agent $i$ and also assigns a likelihood $p^i_k$ with each mode $k$. The colored ellipsoids represent the extent of uncertainty in the predictions.}
\vspace*{-0.7cm}
\label{fig:scenario}
\end{figure}

These challenges highlight the need for robust approaches that explicitly reason about uncertainties in multimodal predictions while actively inferring agent behavior. Consider the scenario in Figure~\ref{fig:scenario}, where the ego vehicle, labelled as $e$ and shown in red, plans to merge into the right lane (indicated by the pink dashed line) while navigating around yellow vehicles. The colored ellipsoids represent uncertain multimodal predictions of other vehicles' trajectories. When the predictor assigns similar likelihoods to multiple potential trajectories for vehicle 1, the ego vehicle risks defaulting to overly conservative behavior. To resolve this ambiguity, the ego vehicle can take two types of actions. The first one is to stay passive, where the ego waits safely while the non-ego vehicles reveal their intention. The other type of action is to take proactive strategic actions that make the non-ego vehicle reveal its intention. The passive strategy results in overly conservative actions that can cause traffic congestion and high braking and accelerating scenarios. Taking careful proactive actions can reveal the intention of non-ego vehicles much faster without being overly conservative. For example, in Figure~\ref{fig:scenario}, if the red ego vehicle brakes passively to wait for vehicle 1 to reveal its intention, it can lead to traffic congestion. However, if the ego vehicle can actively probe vehicle 1 through strategic actions such as accelerating, or steering. If vehicle 1 has an aggressive driver, they may maintain speed or accelerate in response to these probes, while a defensive driver might yield. These interactions reveal crucial information about driver behavior, enabling more informed and efficient trajectory planning. 
% \jd{this paragraph is good but maybe presenting it as a motivating example might be more engaging. eg. highlighting why passive methods are problematic in real-world driving scenarios (e.g - An AV hesitating too long at a busy intersection which is one of your experiments)}

In this paper we present a unified framework that integrates trajectory planning with multimodal predictions and online behavior inference. Our approach explicitly handles prediction uncertainties while enabling active information gathering, demonstrating enhanced safety and robustness across diverse driving scenarios. We demonstrate the effectiveness of our framework in challenging autonomous navigation scenarios using the MetaDrive~\cite{QL-etal:2021_metadrive} simulation environment. Through extensive experiments in merging situations and dense intersections, we demonstrate that our framework significantly outperforms other state-of-the-art planners in safety metrics and task completion rates.

%\noindent\textbf{Related work.}
\subsection{Related work}
The challenge of intention-aware planning in autonomous systems has attracted significant research attention. Early works developed intent recognition models to predict agent behavior based on observed actions~\cite{ZZ-etal:2023,AC-etal:2021,HG-JS-ML-JS:2011}, leading to intent-aware planning frameworks that incorporate these predictions~\cite{CL-SH-HL-JC:2024,HB-etal:2015,SQ-SCZ:2018,RT-LS-MT-DI:2021}. However, these approaches often struggle with the inherent multimodality of human intentions and their limited training data frequently results in prediction errors when encountering novel situations. The branch Model Predictive Control (MPC) ~\cite{YC-etal:2022} and delayed decision-making frameworks \cite{DI-etal:2025_multifuture} offer a method for planning safe trajectories under growing multimodal uncertainty. In the specific context of autonomous navigation, chance-constrained MPC frameworks~\cite{KR-HA-MK:2022,JPA-MB-JCG:2019,SHN-etal:2024,HA-CC-IMM-MK:2021,FL-etal:2021} have been developed to capture diverse driving behaviors and enhance planner robustness. While these approaches represent significant advances, they typically adopt a passive stance, waiting for other agents to reveal their intentions. This passive approach often results in overly conservative behavior that can impede efficient navigation.

Recent approaches to active behavior inference have explored two main directions: dual control mechanisms~\cite{AM:2018,HH-DI-SB-JFF:2024,JK-etal:2024} and information-theoretic probing strategies~\cite{SW-YL-JMD:2023}. While these methods show promise, they rely on joint dynamics between the ego agent and other agents, leading to high-dimensional optimization problems that scale poorly with the number of agents and prediction modes. The computational complexity is particularly problematic when accounting for multimodal intentions, as the state space grows exponentially with each additional mode. This creates a significant gap between theoretical frameworks and practical deployment, where robust handling of multimodal uncertainties is crucial for safe operation. Our work addresses these limitations by developing a computationally efficient framework that explicitly incorporates both prediction uncertainties and multimodal intentions.
% \dgcomment{Need to add multi-future paper}
% \begin{enumerate}
% \item intention-aware planning
% Intent recognition: \cite{ZZ-etal:2023,AC-etal:2021,HG-JS-ML-JS:2011}
% Intent-aware planning: \cite{CL-SH-HL-JC:2024,HB-etal:2015,SQ-SCZ:2018}
% \item chance-constrained MPC, branch MPC:\\
% chance-constrained: \cite{LB-MO-BCW:2011,KR-HA-MK:2022,JPA-MB-JCG:2019}
% Multimodal: \cite{SHN-etal:2024,HA-CC-IMM-MK:2021,FL-etal:2021}
% Branch: \cite{YC-etal:2022}
% \item Active learning, active probing:\\
% Survey on online dual control: \cite{AM:2018}
% Active dual control: \cite{HH-DI-SB-JFF:2024}
% Active learning with Path integral control: \cite{JK-etal:2024}
% Active probing: \cite{SW-YL-JMD:2023}
% \end{enumerate}

\subsection{Contributions}
%\noindent\textbf{Contributions.}  
 We present a unified framework for proactive and robust motion planning that can simultaneously handle multimodal predictions with uncertainties and enable active behavior inference of surrounding agents. Our key contributions are:
\begin{itemize}
        \item A novel \textbf{risk metric} that rigorously quantifies risk by incorporating both multimodal prediction uncertainties and deviations from the ego agent's reference trajectory.
        \item A \textbf{closed-form solution} to risk computation and, proof that the risk is always finite when prediction uncertainties follow a Gaussian mixture distribution, ensuring its tractability.
        \item An \textbf{active probing mechanism} that uses a proactive strategy to infer agent behavior while seamlessly integrating multimodal predictions and their associated uncertainties. We model the interactions between different vehicles directly through the rewards models of the various agents in the scenario, without using their joint dynamics with the ego agent.  Our approach scales efficiently with the number of prediction modes and agents, overcoming key limitations of existing methods. 
\end{itemize}

\section{Problem formulation}

Consider a scenario with $N$ agents and an ego agent $e$. The dynamics of the ego agent are:
\begin{align}
x^e(t+1) = f(x^e(t),u^e(t)), 
\end{align}
where $x^e(t) \in \mathbb{R}^n$ represents the state and $u^e(t) \in \mathbb{R}^m$ represents the control inputs. Our objective is to plan the trajectory of ego agent $e$ using a multimodal predictor, while simultaneously estimating the behavior of the non-ego agents. For each agent $i \in \{1,2,\cdots,N\}$, the ego agent's predictor generates $K$ distinct trajectories $\hat{x}^i_k(t)$, where $k \in \{1,2,\cdots,K\}$, with associated likelihoods $p^i_k$ satisfying $\sum_{k=1}^K p^i_k = 1$. The complete prediction set is
\begin{align*}
s^e = \bigcup_{i=1}^N \bigcup_{k=1}^K \hat{x}^i_k(1:T).
\end{align*}
% Each sample $s^e_l \in s^e$, where $l \in {1,2,\cdots,L}$, represents a specific combination of predicted trajectories. For instance, $s^e_l = {\hat{x}^1_2(1:T), \hat{x}^2_K(1:T), \cdots, \hat{x}^N_3(1:T)}$ denotes a combination where agent 1 follows prediction $k=2$, agent 2 follows prediction $k=K$, and agent $N$ follows prediction $k=3$. 
% For a given set of prediction $s^e$, the ego agent's reward is:
% \begin{align}
% R^e(x^e,u^e,s^e) = \sum_{t=1}^T J^e(x^e(t),u^e(t),s^e(t)).\label{eqn:ego-reward}
% \end{align}
Modern predictors, despite their sophistication, exhibit inherent limitations due to training on finite datasets that inadequately capture rare events and the full diversity of human driving behaviors~\cite{LF-etal:2024_unitraj}.  To address these limitations while improving our understanding of agent behaviors, we aim to estimate the behavior parameters $\phi^i$ for each agent. The ego agent's optimization problem is formulated as:
\begin{align}
\begin{split}
\underset{u^e(1:T)}{\max}~&\alpha_1 R_\text{util}(x^e,u^e) + \alpha_2\mathbb{E}_{s^e \sim \mathbb{P}(S^e)} R_\text{safe}(x^e,u^e,s^e) \\&+ \alpha_3 \sum_{i=1}^N\text{Info}(\phi^i,x^e,u^e),\\
\text{s.t.}\ & x^e(t+1) = f(x^e(t),u^e(t)), \quad t \in [1,T].
\end{split}\label{eqn:ego-problem}
\end{align}
The first objective maximizes a utility-based reward. The second objective relates to safety. The distribution $\mathbb{P}(S^e)$ represents the distribution of all future trajectories of the non-ego agents, and our objective is to maximize the expected safety reward. The third objective maximizes information gained about agent behavior parameters $\phi^i$ through active probing. The weights $\alpha_1, \alpha_2$, and $\alpha_3$ balance the trade-off between utility maximization, safety, and information gain. 
% \di{note: we're assuming the agents behave optimally with regard to \emph{some} objective function, which is not a great assumption, i.e. consider intoxicated, sleeping, distracted agents}
% \dgcomment{I agree with this. I'm currently not sure how to pose the case where the agents are not acting rationally. }

\section{Methodology}

In this section, we present our approach to solving the optimization problem~\eqref{eqn:ego-problem}. We first address each component of the objective function separately: utility maximization for performance, safety considerations for robustness, and information maximization for active behavior inference. We repeat our framework in an MPC fashion to plan the trajectory of the ego agent at every time instance.
\subsection{Utility maximization}

We define the ego agent's utility-based reward $R_\text{util}$ as,
\begin{align}
R_\text{util} &= -\sum_{t=1}^T \|x^e(t)-\bar{x}^e(t)\|^2_Q + \|u^e(t)\|^2_R. \label{eqn:ego-util}
\end{align}
Here, $\bar{x}^e$ represents a reference trajectory for the ego agent. For any vector $y \in R^n$ and positive definite matrix $A \in R^{n \times n}$, the norm $\|y\|^2_A = y^\top A y$. In the utility reward~\ref{eqn:ego-util}, matrices $Q$ and $R$ are positive definite matrices designed to encourage utility-related features such as following a reference trajectory, driving comfort, fuel efficiency, etc. 
\subsection{Safety with risk assessment}
Due to the inherent uncertainty in the evolution of the trajectories of the non-ego agents, we seek to maximize an expected safety reward over all possible future evolutions. Our predictor provides multimodal predictions and likelihoods $(\hat{x}^i_k(1:T),p^i_k)$, where $k \in \{1,\cdots, K\}$ for every vehicle $i \in \{1,\cdots,N\}$. Here, the predicted trajectories are $\hat{x}^i_k(1:T)$ and the associated likelihood is $p^i_k$.
To address the predictor's imperfect predictions, we develop a risk metric for the ego agent with respect to its intended reference trajectory. The ego agent aims to track the reference trajectory as specified in~\eqref{eqn:ego-util}. However, inevitable deviations arise in the future due to dynamic interactions and environmental uncertainties. Therefore, we model the ego's actual trajectory as a sample from a distribution parameterized by the reference trajectory:
\begin{align}
x^e(1:T) &\sim \mathbb{P}^e(1:T), \nonumber\\
\mathbb{E}_{\mathbb{P}^e(1:T)} \left[x^e(1:T)\right] &= \bar{x}^e(1:T). \label{eqn:ego-distribution}
\end{align}
We model the non-ego agent's trajectory using a density mixture parameterized by the predictor outputs as
\begin{align}
x^i(1:T) &\sim \sum_{k=1}^K p^i_k \mathbb{P}^i_k(1:T),\\
\mathbb{E}_{\mathbb{P}^i_k(1:T)}\left[x^i(1:T)\right] &= \hat{x}^i_k(1:T), \label{eqn:GMM-prediction}
\end{align}
where $\mathbb{P}^i_k$ represents the distribution for the $k$-th mode with mean $\hat{x}^i_k$ and mixing weight $p^i_k$. Given these probabilistic representations of the trajectories of the agents, we define the risk associated with following reference trajectory $\bar{x}^e$ relative to each predicted mode of agent $i$ as:
\begin{align}
r^{i}_k(t) &= p^i_k \left(1 + e^{-\alpha \mathcal{W}(\mathbb{P}^e(t),\mathbb{P}^i_k(t))}\right), \label{eqn:risk}
\end{align}
where $\mathcal{W}(\mathbb{P}^e,\mathbb{P}^i_k)$ denotes the 2-Wasserstein distance~\cite{GP-MC:2019_optimaltransport} between distributions $\mathbb{P}^e$ and $\mathbb{P}^i_k$, and $\alpha > 0$ scales the sensitivity to distributional differences. 
% The 2-Wasserstein distance between any two distributions $\mathbb{P}$ and $\mathbb{Q}$ which are supported on $\mathbb{R}^n$ is formally given by
% \begin{align}
% \mathcal{W}^2(\mathbb{P},\mathbb{Q}) &=\inf_{\pi \in \beta(\mathbb{P},\mathbb{Q})} \int_{\mathbb{R}^n \times \mathbb{R}^n} \left|\left| x^p - x^q \right|\right| \mathrm{d}\pi(x^p,x^q),\label{eqn:wasserstein}
% \end{align}
% where $\pi$ is a joint distribution between $\mathbb{P}$ and $\mathbb{Q}$, and $\beta$ is the set of all possible joint distributions. 
\begin{remark}\textbf{(Probabilistic risk metric)}While CVaR-based risk~\cite{AM-MP:2019_axiomaticrisk} could handle Gaussian mixtures tractably, our Wasserstein-based risk metric directly incorporates prediction uncertainties and provides closed-form computation. Unlike CVaR which focuses on tail violation, our metric captures distributional proximity weighted by prediction confidence. We select the 2-Wasserstein distance $\mathcal{W}^2$ as our distributional metric for several mathematical reasons. Unlike divergence measures such as the Kullback-Leibler divergence or Gaussian overlap~\cite{FMT-DI-JSB-SB:2023}, the Wasserstein distance satisfies all properties of a true metric including the triangle inequality. This is particularly valuable when comparing distributions $P^e$ and $P_k^i$ with different supports, where KL divergence would be infinity, hence computationally intractable. The Wasserstein distance has a natural interpretation as the minimum cost of moving one distribution into another, capturing differences in both shape and location. 
\end{remark}

In the following result, we give a closed-form expression to compute the risk and ensure its tractability.
\begin{lemma} \textbf{(Analytical tractability of risk)}
Let each mode of prediction $\mathbb{P}^i_k(1:T)$ be a Gaussian distribution such that $\mathbb{P}^i_k(1:T) = \mathcal{N}(\hat{x}^i(1:T),C^i_k(1:T))$ where $C^i_k(1:T)$ is the trajectory of covariance matrices. Similarly, let the probability distribution of the trajectory of the ego agent $\mathbb{P}^e(1:T)$ also be a Gaussian distribution such that $\mathbb{P}^e(1:T) = \mathcal{N}(\bar{x}^e(1:T),C^e(1:T))$. Then, risk $r^i_k$ is finite. 
\end{lemma}

\noindent\emph{Proof.} We first assume that the state space is bounded, which is a fair assumption because the positions, velocities, and control inputs are always bounded in realistic planning scenarios. The covariances $C^i_k$ and $C^e$ are finite positive definite matrices. The 2-Wasserstein distance between the ego's distribution $P^e$ and the traffic agent $i$'s distribution $k-$th mode $P^i_k$ is given by, 
\begin{align}
\mathcal{W}^2\left(\mathbb{P}^e(t),\mathbb{P}^i_k(t)\right) &= \left|\left|\bar{x}^e(t) - \hat{x}^i_k\right|\right| \\ 
&- 2 ~ \mathrm{Tr}\Big[C^e(t) + C^i_k(t)  \nonumber  \\ &+ \Big({C^e(t)}^{\frac{1}{2}}  
C^i_k(t)  {C^e(t)}^{\frac{1}{2}}\Big)^{\frac{1}{2}} \Big]. \nonumber
\end{align}
The above distance metric between distributions $\mathbb{P}^e(t)$ and $\mathbb{P}^i_k(t)$ is clearly finite, which renders the risk finite. \hfill $\blacksquare$

\begin{figure}[tbh]
\begin{multicols}{2}
% \hspace*{1cm}
\begin{tikzpicture}
  \node (img1)  {\includegraphics[width=0.170\textwidth]{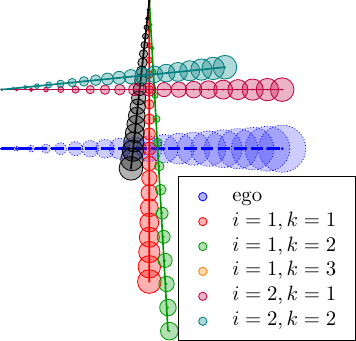}};
  \node[below of= img1, node distance=0cm, xshift = 0.0cm, yshift=-2.3cm,font=\color{black}]  {(a)};
% \node(i1) at (0,1.7) {$i=1$};
\end{tikzpicture}\columnbreak
% \hspace*{0.1cm}
\begin{tikzpicture}
  \node (img2)  {\includegraphics[width=0.250\textwidth]{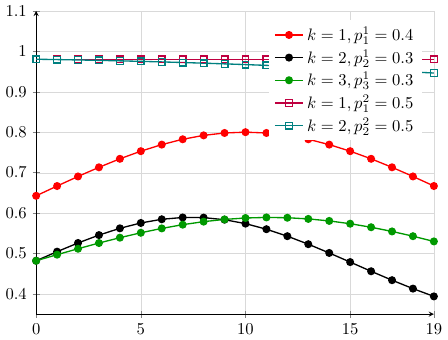}};
  \node[above of= img2, node distance=0cm, xshift = 0.0cm, yshift=-2.6cm,font=\color{black}]  { (b)};
  \node[left of= img2, node distance=0cm, rotate=90, anchor=center,yshift=2.5cm,font=\color{black}] {Risk $r^{i}_k(t)$};
  \node[above of= img2, node distance=0cm, xshift = 0.0cm, yshift=-2cm,font=\color{black}]  { Time $t$};
\end{tikzpicture}
\end{multicols}
\vspace*{-1.92em}
\caption{Risk assessment with multiple agent trajectories. (a) Predicted trajectories for an ego agent (blue reference path) interacting with two other agents. Agent 1's predicted paths (black, red, green) run top to bottom with circles indicating prediction uncertainty. Agent 2's predicted paths (purple, teal) run left to right. (b) Risk $r_k^i(t)$ over time for each predicted trajectory. Agent 1's red prediction shows the highest risk around $t=10$ due to higher probability and overlap with the ego's trajectory, but eventually reduces as their trajectories diverge. Agent 2's parallel prediction $k=1$ maintains high risk while its diverging prediction $k=2$ shows decreasing risk.}
\label{fig:risk-wasserstein}
\end{figure}
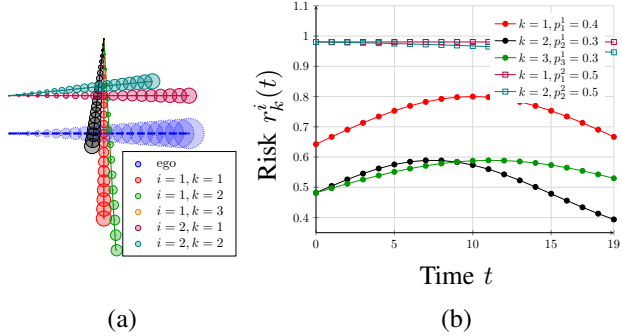

\begin{example}(\textbf{Risk assessment with $N=2$ agents}) We illustrate our risk assessment framework in Figure~\ref{fig:risk-wasserstein}. In Figure~\ref{fig:risk-wasserstein}(a), the ego agent's planned trajectory, depicted in blue, goes from left to right. There are $N=2$ vehicles in the scene. For agent $1$, there are three predicted trajectories that go vertically from the top to bottom. They are represented in red, black, and green, with likelihoods $p^1_1 = 0.4$, $p^1_2 = 0.3$, $p^1_3 = 0.3$, respectively. For agent $2$, there are two predicted trajectories that run from left to right, similar to the ego agent. They are represented in purple and teal, with equal likelihoods $p^2_1=0.5$ and $p^2_2=0.5$. It is important to note that the covariances of the predictions grow over time representing the fact that our predictions get poorer further into the future. In Figure~\ref{fig:risk-wasserstein}(b), we plot the risk $r^{i}_k(t)$ versus time for both the agents and their corresponding predicted trajectories. For agent 1, it is clear that prediction $k=1$, which overlaps with the ego agent's trajectory at time $t=10$, has the highest risk. For predictions 2 and 3, although there is an overlap, the ego is further away at any given time compared to prediction 1. Agent 2 runs parallel to the ego agent, and their proximity results in higher risk. For prediction $k=1$ which is exactly parallel to the ego, the risk $r^2_1(t)$ is close to 1 and is flat throughout, which implies that there is consistently high risk. However, for prediction $k=2$, the agent deviates away from the ego, which results in decreasing risk over time.
\end{example}
% A practical aspect is the computational complexity of computing the risk. The computational complexity for computing the risk of the order $\mathcal{O}(n^3)$, where $n$ is the dimension of the state space. The main computation complexity arises from the matrix multiplication and inversion operations involved in computing the Wasserstein metric.
% \di{If I'm understanding the risk correctly, it's based on overlap. Does this mean that two orthogonal trajectories that overlap at a particular time instant might have a lower risk than two parallel trajectories which never overlap, but are close together for the entire time horizon?}
% \dgcomment{It is not completely based on overlap. The 2-wasserstein distance captures the amount of effort it takes to transform the ego's density $\mathbb{P}^e$ to the non-ego's density $\mathbb{P}^i_k$. This depends on the trajectory of means and covariances. For the sake of clarity, let's assume the covariance for both the parallel and perpendicular scenarios are the same. For the perpendicular scenario, the risk will look similar to 2(b) where it is initially low and then as the non-ego agent gets closer the risk increases. For the parallel scenario, the risk would be consistently high. 
% I added a parallel scenario with corresponding risks to illustrate risk better.
% }

Given the probabilistic risk metric, we ideally need to impose safety constraints as
\begin{align*}
  % w^{i}_k(t) &= x^e(t) - \hat{x}^i_k(t) \\
  \|x^e(t) - \hat{x}^i_k(t)\|_{{C^i_k}^{-1}(t)} &\geq L~r^i_k(t)
  % {w^{i}_k}^\top(t) {C^i}^{-1}_k v^{i}_k(t) &\geq L ~ r^{i}_k(t),
\end{align*}
where $L$ is a predetermined safe distance between the agent $i$ and the ego. Here the weighted norm is with respect to the inverse of the covariance ${C^i_k}$, which ensures that for predictions with higher uncertainty, we need a higher safe distance between the ego and the agent $i$. Note that the safe distance constraint needs to be satisfied according to the risk $r^i_k(t)$, meaning that for cases with lower risk, the constraint is easier to satisfy.  % \di{I think we're overloading $v$, it's used for velocity in Eq. 16 and after. $v$ for velocity is more common, so better to change it here for the safety constraint. Maybe $d$ for displacement, not the best since we use $d\pi$ in Eq 10, but I don't think we use a standalone $d$ anywhere. Maybe you have a better idea. Whatever you decide, you should also define in with a verbal explanation beyond the equation definition. also equation definitions typically use a different equation sign. either 3 bars or an equal sign with a triangle over.} 
However, the constraint is non-convex and difficult to handle computationally. To make the safety constraint tractable, we relax the above constraint to a soft constraint in the objective function as
\begin{align}
q^{i}_k(t)  &= \|x^e(t) - \hat{x}^i_k(t)\|_{{C^i_k}^{-1}(t)} - L~r^i_k(t), \nonumber \\
R_\text{safe} &= \sum_{t=1}^T \sum_{i=1}^N \sum_{k=1}^K \log(1 + e^{-\beta q^{i}_k(t)}).\label{eqn:ego-safety-cost}
\end{align}
Here, the variable $q^{i}_k(t)$ considers the gap between weighted distance $\|x^e(t) - \hat{x}^i_k(t)\|_{{C^i_k}^{-1}(t)}$ and the safe distance prescribed by the risk as $L r^i_k(t)$. The log-exp function behaves like a soft barrier function where $\beta$ is a factor that controls the sharpness of the barrier. Higher values of $\beta$ denote a conservative barrier where even large gaps $q^i_k(t)$ will lead to conservative actions, and vice versa. Therefore, the safety cost $R_\text{safe}$ is a smooth function which makes the problem computationally tractable using standard gradient-based optimization tools. 

% \di{$R_{safety}$ from eq.4 should have been a result of this section. where is it?}
% \dgcomment{Equation (13) has $R_{safety}$. I had forgotten to change some notation earlier.}

\subsection{Active probing for behavior inference }

Our third objective is to infer the behavior of other agents by taking proactive probing actions. At each time instance, the ego assumes that agent $i$ uses control actions $u^i(t)$ that maximize the reward:
\begin{align}
R^i(x^i,u^i,s^i) = \sum_{t=1}^T {\phi^i}^\top J^i\left(x^i(t), u^i(t), s^i(t)\right). \label{eqn:agent-reward-model}
\end{align}
Here, the reward $R^i$ has an objective vector $J^i$ with contextual features that capture various aspects of the scenario such as inter-agent safety measures, deviation from reference trajectories, control effort penalty, comfort metrics, etc~\cite{DB-GA-VK:2024}. The behavior parameter $\phi$ weights these features according to agent $i$'s preferences. We equip the ego agent with the objective vector $J^i$ and an initial guess of the behavior parameter $\phi^i$. Note that the objectives depend on the state and control of the respective agent $i$, and also on $s^i$ . Here $s^i$ is the set that encompasses predicted state trajectories for all agents from the perspective of the agent $i$.  The interactions between agent $i$ and any other agent $j$ are captured through $s^i$. For example, if the trajectories of $i$ and $j$ are at close proximity or intersect, the reward associated with safety is low, and vice versa.  Given the reward model~\eqref{eqn:agent-reward-model}, at each time instance, agent $i$ is assumed to compute its control sequence by solving:
\begin{align*}
u^i(1:T) = \argmax_{u^i} ~ R^i(x^i,u^i,s^i).
\end{align*}
Agent $i$ can implement this reward maximization through various frameworks such as MPC, reinforcement learning, or imitation learning. We assume agents are rational with respect to their objectives but non-adversarial. They optimize their individual rewards while avoiding intentionally dangerous behaviors such as collisions or deliberate disruptions.

We describe our methodology to improve our estimates of the behavior parameters $\phi$ through active probing. We note that the likelihood of every mode $k$ of agent $i$ is truly dependent on the states and actions of the ego agent. To capture this, we modify the Boltzmann model of decision-making~\cite{BDZ-etal:2008,JFF-etal:2018,SW-YL-JMD:2023} informed by the predicted likelihood $p^i_k$. The modified Boltzmann model is
\begin{align}
\hat{p}^i_k(x^e,u^e)= \frac{p^i_k e^{R^i(x^i_k,u^i_k, s^i(x^e,u^e))}}{\sum_{j=1}^K p^i_j e^{R^i(x^i_j,u^i_j,s^i(x^e,u^e))}}. \label{eqn:boltzmann}
\end{align}
The above model has an exponential weight on the reward $R^i$ of each trajectory $k$ of agent $i$. It is important to note that the rewards of each trajectory $k$ depends on the actions of the ego agent. The reward $R^i$ is a function of the interactions between the different agents through the predicted state trajectories $s^i$. Since the state trajectory of the ego agent influences the reward of agent $i$, we explicitly use $s^i(x^e,u^e)$ in~\eqref{eqn:boltzmann}. The interaction between the agent $i$ and the ego agent affects the reward associated with each mode $k$~\eqref{eqn:agent-reward-model}. Due to the rationality of the agents, higher rewards result in higher likelihoods of future trajectories of agent $i$. The modification to the Boltzmann model is in the scaling factor $p^i_k$ which is available to us from the predictor, but not entirely reliable. 

To estimate the reward parameters $\phi^i$, we construct a belief distribution using an initial guess $\hat{\phi^i}$ as $b(\hat{\phi^i}) = \mathcal{N}(\hat{\phi^i},\Sigma)$. Here,
% \begin{align*}
%   b(\hat{\phi^i}) = \mathcal{N}(\hat{\phi^i},\Sigma),
% \end{align*}
$\Sigma$ is a given positive definite covariance matrix. If the true future trajectory is $k$, we can update the belief distribution using the Bayes' rule as $ b'_k(\hat{\phi^i},x^e,u^e) \propto b(\hat{\phi^i}) \hat{p}^i_k(x^e,u^e).$
% \begin{align*}
%   b'_k(\hat{\phi^i}) \propto b(\hat{\phi^i}) \hat{p}^i_k.
% \end{align*}
To improve our beliefs as best as possible, we maximize the information radius between $b(\hat{\phi^i})$ and $b'_k(\hat{\phi^i})$, where the information radius is defined as the KL divergence between the two distributions~\cite{MF-etal:2014,SW-YL-JMD:2023}. The KL divergence between any two distributions $P$ and $Q$ is 
\begin{align*}
\text{KL}(P||Q) &= \int P(x) \log \left(\frac{P(x)}{Q(x)}\right) \mathrm{d}x.
\end{align*}
Practically, information gain is a quantification of the difference in the shape of the belief distribution expressed through the KL divergence. Since the likelihood $\hat{p}^i_k$ is a function of the ego agent's actions, we are taking active actions to improve our belief of the behavior of the other agents. However, the best action to take to maximize the KL divergence could lead to reduced safety or jerky motion in navigation scenarios. Therefore, it is essential to regulate the extent of probing in risky situations. We use a simple threshold-based cutoff to limit probing in risky situations. Particularly, if $ r^i_k(t) > \tau$, where $\tau$ is a threshold, we do not probe the agent $i$ for mode $k$. This ensures that the ego agent does not take risky actions to probe the other agents. We define the information gain for agent $i$'s $k$-th mode as
\begin{align}\label{eqn:info-gain-KL}
\text{Info}^i_k(\phi^i,x^e,u^e) &= \begin{cases}0, & r^i_k(t) > \tau, \\[0.3cm] \text{KL}\left(b\Big|\Big|b'_k\right), & \text{otherwise}.
\end{cases}
\end{align}

This notion of information gain depicts the change in the shape of our belief distribution $b(\hat{\phi^i})$ when we update it with the likelihood $\hat{p}^i_k(x^e,u^e)$. If the risk for mode $k$ is high, we gain no information. We drop the arguments of $b$ and $b'_k$ only for readability. The total information gained for agent $i$ is
\begin{align}
\text{Info}(\phi^i,x^e,u^e) = \frac{1}{K}\sum_{k=1}^K\text{Info}^i_k(\hat{\phi^i},x^e,u^e)\label{eqn:info-gained-i}
\end{align}

The information gain is a measure of how much we learn about the behavior of agent $i$ by probing it with different actions. We use a particle filter to implement the information maximization objective. We start with $M$ particles sampled from the prior belief $b(\hat{\phi^i})$. With $K$ different trajectories, the updated belief $b'_k(\hat{\phi^i},x^e)$ can be computed for each particle through the modified Boltzmann model~\eqref{eqn:boltzmann}. For every mode $k$, the information gain is computed using $M$ particles as
\begin{align*}
  \text{KL}\left(b||b'_k\right) \approx \frac{1}{M}\sum_{m=1}^M b(\hat{\phi^{i}}(m)) \log\frac{b(\hat{\phi^{i}}(m))}{b'_k(\hat{\phi^i},x^e)}.
\end{align*}
Here, $b(\hat{\phi^{i}}(m))$ is the likelihood of the $m^{\text{th}}$ particle based on the Gaussian distribution $b(\hat{\phi^i})$. 

We provide an implementation of our framework in Algorithm~\ref{algo:framework}. We repeat Algorithm~\ref{algo:framework} at each time instance in an MPC fashion to plan the trajectory of the ego agent. 

\begin{algorithm}
  \SetAlgoLined
  \caption{Trajectory planning with multimodal predictions and active probing}
  \label{algo:framework}
  \KwData{Multimodal predictions $\hat{x}^i_k(0:T)$, corresponding likelihoods $p^i_k$, for $k=1,\cdots, K$;  Initial belief $\hat{\phi^i}$; Risk threshold $\tau$; Initial state $x^e(0)$; Reference trajectory $\bar{x}^e(0:T)$;  }
  \KwResult{ $x^e(0:T), u^e(0:T)$ }
  \For{$i=1$ to $N$}{
  \For{$k=1$ to $K$}{
     Compute risk $r^i_k(1:T)$ ~\eqref{eqn:risk}\\
       \If {$r^i_k(t) > \tau$}{
    $\text{Info}(\hat{\phi^i}) = 0.$ \\
   }     
   \Else{
    Compute likelihoods $\hat{p}^i_k$~\eqref{eqn:boltzmann}\\
    Compute Info$(\phi^,x^e,u^e)$~\eqref{eqn:info-gained-i}
    }
  }

  }
  \While{not converged}{
    Solve problem~\eqref{eqn:ego-problem}, with costs~\eqref{eqn:ego-util},~\eqref{eqn:ego-safety-cost},~\eqref{eqn:info-gain-KL} subject to ego agent's dynamics.
  }
  \end{algorithm}

\section{Numerical Analysis}\label{sect:experiments}
We validate our algorithm in autonomous navigation scenarios using the MetaDrive\cite{QL-etal:2021_metadrive} simulation environment, chosen for its realistic vehicle dynamics and diverse traffic scenarios. Our evaluation spans two distinct scenarios: a merging scenario with a lane change maneuver, and an intersection crossing scenario without signal lights or stop signs. We compare our approach against \cite{KR-HA-MK:2022}, a state-of-the-art chance-constrained planner utilizing multimodal predictions without probing capabilities. This section details our simulation setup, experimental results, and computational performance analysis. We model all vehicles, including the ego vehicle, using a bicycle model with state vector $x = [x, y, \theta, v]^\top$, where $(x,y)$ denotes the vehicle's position, $\theta$ represents the heading angle, and $v$ is the velocity. The dynamics are:
\begin{align}
  \begin{split}
    \dot{x} &= v \cos(\theta), \quad \dot{y} = v \sin(\theta), \\ \dot{\theta} &= \omega, \quad \dot{v} = a, \quad 
  \end{split}
\end{align}
where $a$ and $\omega$ are the acceleration and steering inputs, respectively. For numerical implementation, we discretize these dynamics using Euler integration with a 0.1$s$ time step.

The ego vehicle employs the Gaussian Lane Keeping (GLK) predictor~\cite{DI-PG-XL-SB:2024_GLK}, which generates $K=3$ distinct predictions by combining constant velocity and lane-snapping behaviors through Bayesian inference. For each agent $i$, GLK provides trajectory predictions $\hat{x}^i_k(1:T)$, associated error covariances $C^i_k(1:T)$, and prediction likelihoods $p^i_k$. The ego vehicle needs to estimate the parameters $\phi^i$ of each agent which defines their behavior. To this end, we assume that the ego agent uses a reward model for agent $i$  inspired from the Intelligent Driver Model~\cite{MT-AH-DH:2000} that captures key driving behaviors
\begin{align}
  R^i(x^i,u^i,s^i) = \sum_{t=1}^T \begin{bmatrix}\phi^i_1 \\ \phi^i_2 \\ \phi^i_3 \end{bmatrix}^\top \begin{bmatrix} -\|v^i(t) - \bar{v} \| \\ \|x^i(t) - x^e(t)\| \\ \|x^i(t) - \bar{x}^i(t)\| \end{bmatrix}.\label{eqn:ego-perspective-agent-reward-model}
\end{align}
Here, $\phi^i = [\phi^i_1 \ \phi_2 \ \phi_3]$ weights three key behaviors: velocity matching, maintaining safe distances, and lane positioning. While this model approximates expected agent behavior, we note that the true agent state prediction $s^i$ is unobservable, leading us to focus on estimating the safety parameter relative to the ego agent. To evaluate robustness, we vary the actual controllers of other agents across scenarios. Table~\ref{tab:params} summarizes the key simulation parameters maintained across all experiments.

\begin{table}
  \centering
  \caption{Simulation parameters}\label{tab:params}
  \begin{tabular}{|| c | c | c ||}
    \hline
    \textbf{Parameter} & \textbf{Description} & \textbf{Value} \\ \hline
    $T$ & Time horizon & $2.5s$ \\  
    $K$ & No. of predictions per agent & 3  \\  
    $\alpha_1$ & Utility weight & 0.9  \\
    $\alpha_2$ & Safety weight & 0.9  \\
    $\alpha_3$ & Info weight &0.1 \\
    $L$ & Safe distance & $4m$ \\ 
    $\beta$ & Safety barrier & 0.02  \\
    $\tau$ & Risk threshold & 5  \\
    \hline
\end{tabular}
\vspace*{-0.4cm}
\end{table}

\subsection{Lane change scenario with mixed driving behaviors}
We evaluate our algorithm in a lane-change scenario where the ego vehicle must merge from its initial lane into the middle lane while interacting with $N=3$ other vehicles as seen in Figure~\ref{fig:merging-scene_aggr_def}. To test our algorithm's ability to identify and exploit interaction opportunities, we assign either aggressive or defensive behaviors to the other vehicles. Aggressive vehicles prioritize maintaining high velocities, while defensive vehicles tend to yield to maintain safe distances. The controllers for these behaviors are generated by optimizing the reward function:
\begin{align}
  R^i(x^i,u^i,s^i) = \sum_{t=1}^T \begin{bmatrix}\phi_1 \\ \phi_2 \\ \phi_3\end{bmatrix}^\top \begin{bmatrix} -\|v^i(t) - \bar{v} \| \\ \sum\limits_{j \neq i}\|x^i(t) - \hat{x}^j(t)\| \\ \|x^i(t) - \bar{x}^i(t)\| \end{bmatrix},  
\end{align}
subject to the bicycle dynamics constraints. The behavioral characteristics emerge from different weightings of the reward components: aggressive agents emphasize velocity matching ($\phi_1 = 0.5$, $\phi_2 = 0.25$, $\phi_3 = 0.25$), while defensive agents prioritize maintaining safe distances ($\phi_1 = 0.2$, $\phi_2 = 0.6$, $\phi_3 = 0.2$). We examine three distinct scenarios, all starting from the same initial configuration as in Figure~\ref{fig:merging-scene_aggr_def}(a).
\subsubsection{Defensive trailing vehicle} When the last vehicle in the target lane (gray vehicle in Figure~\ref{fig:merging-scene_aggr_def}(b)) exhibits defensive behavior while others remain aggressive, our algorithm successfully probes and identifies this vehicle's tendency to yield, enabling a safe merge in front of it.

\subsubsection{Defensive middle vehicle} 
\begin{figure}[tbh]
  \begin{multicols}{2}
  % \hspace*{1cm}
  \begin{tikzpicture}
    \node (img1)  {\includegraphics[width=0.20\textwidth]{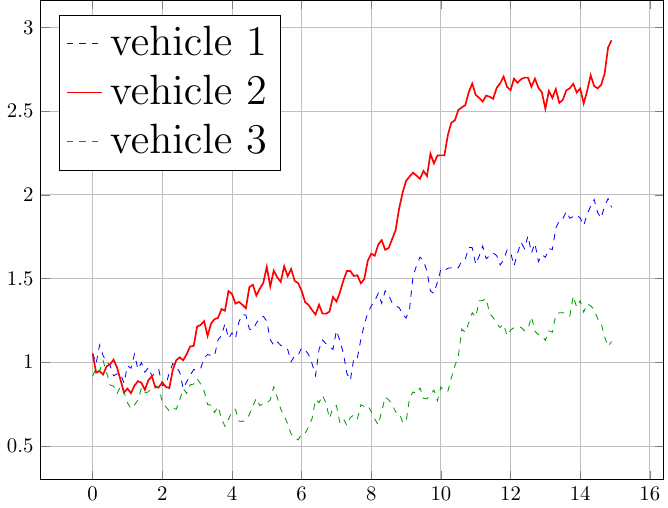}};
    \node[above of= img1, node distance=0cm, xshift = 0.0cm, yshift=-2.2cm,font=\color{black}]  { (a) Scene A.2};
    \node[left of= img1, node distance=0cm, rotate=90, anchor=center,yshift=2.0cm,font=\color{black}] {$\hat{\phi}_2(t)$};
    \node[above of= img1, node distance=0cm, xshift = 0.0cm, yshift=-1.7cm,font=\color{black}]  { Time $t$};
  \end{tikzpicture}\columnbreak
  \hspace*{0.1cm}
  \begin{tikzpicture}
    \node (img1)  {\includegraphics[width=0.20\textwidth]{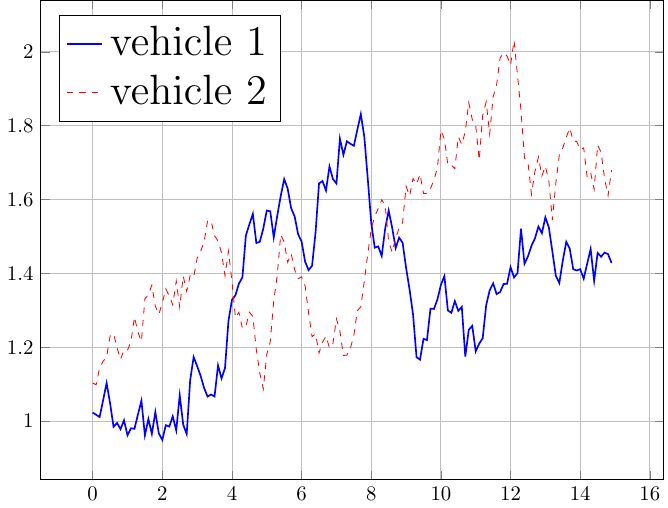}};
    \node[above of= img1, node distance=0cm, xshift = 0.0cm, yshift=-2.2cm,font=\color{black}]  { (b) Scene B.2};
    % \node[left of= img1, node distance=0cm, rotate=90, anchor=center,yshift=2.5cm,font=\color{black}] {Risk $\hat{\phi}_2(t)$};
    \node[above of= img1, node distance=0cm, xshift = 0.0cm, yshift=-1.7cm,font=\color{black}]  { Time $t$};
  \end{tikzpicture}
  \end{multicols}
  \vspace*{-2em}
  \caption{Particle filter estimation of safety weights over time. (a) Scenario 2: Higher estimated safety weight for vehicle 2 confirms its defensive behavior. (b) Time-varying scenario: Estimation tracks vehicle 1's transition from defensive to aggressive behavior at $t=7s$, with slight detection delay.}
  \label{fig:merging-scene-PF-estimation}
  \end{figure}
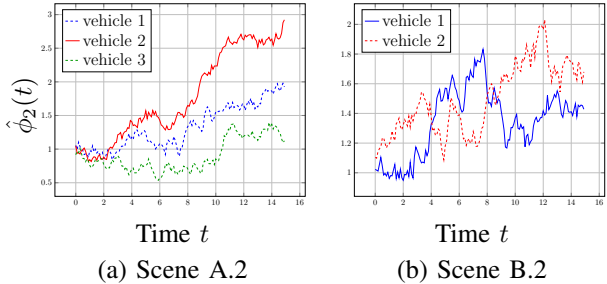

With the middle vehicle is defensive and others aggressive, the ego vehicle successfully identifies and leverages this behavior to execute a merge ahead of the defensive agent as seen in Figure~\ref{fig:merging-scene_aggr_def}(c). For this scenario, we also analyze the particle filter's estimation of behavioral parameter $\hat{\phi}^i_2$. Figure~\ref{fig:merging-scene-PF-estimation}(a) shows the estimated weights on safety over time for all three vehicles. The particle filter correctly identifies the defensive behavior of vehicle 2, as shown by its higher estimated $\hat{\phi}_2$ value, enabling the ego vehicle to plan its merge maneuver accordingly. 

\subsubsection{All aggressive vehicles} 
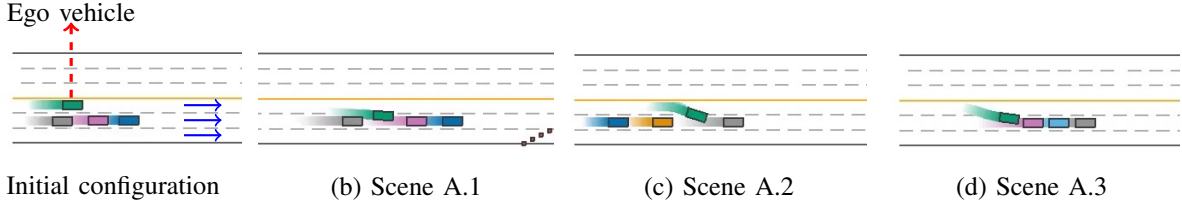
\begin{figure*}
  \begin{multicols}{4}
    % \hspace*{0.5cm}
    \begin{tikzpicture}
      \node (img1)  {\includegraphics[width=0.2200\textwidth]{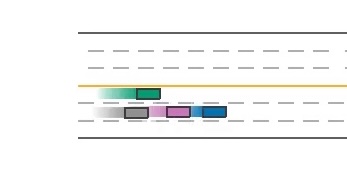}};
      \draw[->, thick, blue] (1.2,-0.1) -- (1.7,-0.1);
      \draw[->, thick, blue] (1.2,-0.3) -- (1.7,-0.3);
      \draw[->, thick, blue] (1.2,-0.5) -- (1.7,-0.5);

      \draw[->, dashed, very thick, red] (-0.3,0) -- (-0.3,1);
      \node at (-0.3,1.1) {Ego vehicle};
      \node[below of= img1, node distance=0cm, yshift=-1.206cm,font=\color{black}]  { (a) Initial configuration};
    \end{tikzpicture}\columnbreak
    % \hspace*{0.7cm}
    \begin{tikzpicture}
      \node (img1)  {\includegraphics[width=0.2200\textwidth]{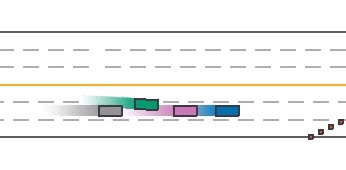}};
      \node[below of= img1, node distance=0cm, yshift=-1.20cm,font=\color{black}]  {  {(b) Scene A.1}};
    \end{tikzpicture}\columnbreak
    % \hspace*{1cm}
    \begin{tikzpicture}
      \node (img1)  {\includegraphics[width=0.2200\textwidth]{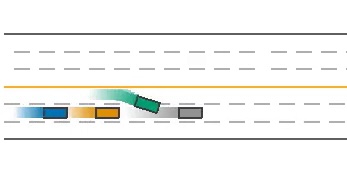}};
      \node[below of= img1, node distance=0cm, yshift=-1.2cm,font=\color{black}]  {  (c) Scene A.2};
    \end{tikzpicture}\columnbreak
    \begin{tikzpicture}
      \node (img1)  {\includegraphics[width=0.2200\textwidth]{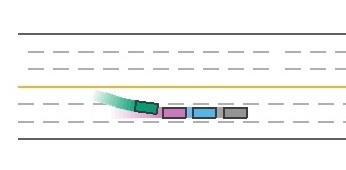}};
      \node[below of= img1, node distance=0cm, yshift=-1.2cm,font=\color{black}]  {(d) Scene A.3};
    \end{tikzpicture}    
    \end{multicols}
    \vspace*{-2em}
    \caption{Lane change scenarios demonstrating our algorithm's adaptation to different vehicle behaviors. The ego vehicle (green rectangle) starts in the topmost lane and aims to merge into the middle lane. (a) Initial configuration of all vehicles. The trail of respective vehicle color denotes history. Driving direction is from left to right. (b) Scenario 1: The last vehicle (gray) exhibits defensive behavior, allowing the ego vehicle to probe and create a merging gap. (c) Scenario 2: The middle vehicle is defensive while others are aggressive; the ego vehicle identifies and exploits this behavior to merge ahead. (d) Scenario 3: All vehicles exhibit aggressive behavior, leading the ego vehicle to execute a conservative merge behind the last vehicle.}
    \label{fig:merging-scene_aggr_def}

\end{figure*}

When all vehicles exhibit aggressive behavior, our algorithm correctly identifies the lack of yielding opportunities and safely executes a conservative merge behind the last vehicle as seen in Figure~\ref{fig:merging-scene_aggr_def}(d).

These scenarios demonstrate our algorithm's capability to not only identify different behavioral patterns through probing but also to adapt its merging strategy accordingly.

% \textcolor{red}{\jd{clarify fig 4b}}

\subsection{Agents with time-varying behavior}
We evaluate our algorithm's robustness to time-varying behaviors with $N=2$ vehicles in the same lane-change scenario. Trailing vehicle $i=1$ transitions from defensive to aggressive behavior, while vehicle $i=2$ remains consistently aggressive. We use the same reward function and weight parameters as the previous scenario to define these behaviors.

\subsubsection{Behavior switch at $t=2$} In this scenario, vehicle $i=1$ switches to aggressive behavior at $t=2s$, too early for the ego vehicle to merge which is evident in Figure~\ref{fig:varying-behavior}(a). The ego is only partially able to enter the target lane. 
\subsubsection{Behavior switch at $t=7$} 
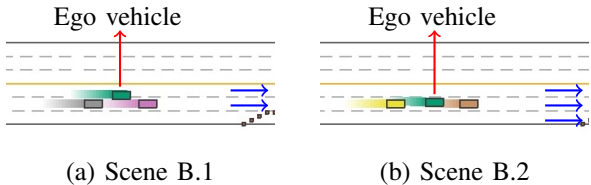
\begin{figure}[tbh]
  \begin{multicols}{2}
  % \hspace*{1cm}
  \begin{tikzpicture}
    \node (img1)  {\includegraphics[width=0.20\textwidth]{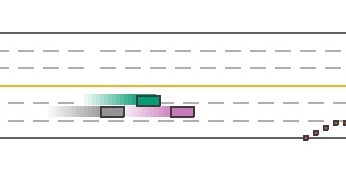}};
    \node[above of= img1, node distance=0cm, xshift = 0.0cm, yshift=-1.2cm,font=\color{black}]  { (a) Scene B.1};
      \draw[->, thick, blue] (1.2,-0.1) -- (1.7,-0.1);
      \draw[->, thick, blue] (1.2,-0.3) -- (1.7,-0.3);
      % \draw[->, thick, blue] (1.2,-0.5) -- (1.7,-0.5);

      \draw[->, thick, red] (-0.27,-0.05) -- (-0.27,0.7);
      \node at (-0.3,0.85) {Ego vehicle};
  \end{tikzpicture}\columnbreak
  \hspace*{0.3cm}
  \begin{tikzpicture}
    \node (img1)  {\includegraphics[width=0.20\textwidth]{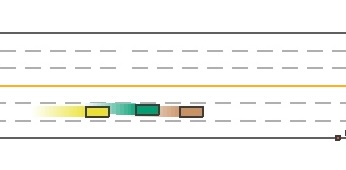}};
      \draw[->, thick, blue] (1.2,-0.1) -- (1.7,-0.1);
      \draw[->, thick, blue] (1.2,-0.3) -- (1.7,-0.3);
      \draw[->, thick, blue] (1.2,-0.5) -- (1.7,-0.5);

      \draw[->, thick, red] (-0.27,-0.15) -- (-0.27,0.7);
      \node at (-0.3,0.85) {Ego vehicle};   
    \node[above of= img1, node distance=0cm, xshift = 0.0cm, yshift=-1.2cm,font=\color{black}]  { (b) Scene B.2};
  \end{tikzpicture}
  \end{multicols}
  \vspace*{-2em}
  \caption{Lane change experiments with time-varying agent behavior. The ego vehicle (green) attempts to merge into the middle lane ahead of vehicle 1, which switches from defensive to aggressive behavior. (a) Early behavior switch at $t=2s$ prevents successful merging. (b) Later switch at $t=7s$ allows sufficient time for the ego vehicle to merge.}
  \vspace*{-0.4cm}
  \label{fig:varying-behavior}
  \end{figure}

In this scenario, the behavior switch occurs at $t=7s$. The ego vehicle successfully probes and exploits the initial defensive period to complete its merge which can be seen in Figure~\ref{fig:varying-behavior}(b). Figure~\ref{fig:merging-scene-PF-estimation}(b) shows the estimated safety weights $\phi^i_2$ for both vehicles in the second experiment. We observe the particle filter detecting the behavioral change with a delay at $t=7.7s$, allowing the ego vehicle to adapt its merging strategy accordingly.

\subsection{Monte Carlo simulations - lane change scenario}
We compare our active probing algorithm against a state-of-the-art chance-constrained MPC planner (CC-MPC) that uses multimodal predictions without probing~\cite{KR-HA-MK:2022}, and against our algorithm with probing disabled ($\alpha_2=0$). We conduct 200 Monte Carlo episodes of the merging scenario with $N=3$ vehicles.  We randomize the spawn positions and behavior parameters in our Monte Carlo experiments. For each episode, we maintain the spawn positions across the different algorithms for fairness. To demonstrate the robustness of our algorithm to a wide spectrum of behaviors, we sample behavior parameters $\phi^i$ from a Gaussian with mean $[0.5 \ 0.3 \ 0.3]^\top$ for aggressive vehicles and $[0.2 \ 0.6 \ 0.2]^\top$ for defensive vehicles, and maintain it across the algorithms. 

Table~\ref{tab:montecarlo-merging} compares the algorithms' performance metrics. We define a successful episode as one where the ego vehicle's entire body is within the limits of the target lane. Our active probing algorithm achieves higher success rates and faster merging times. Further, our algorithm has faster merging times resulting from probing. This arises from the probing actions that reveal the intentions of the other vehicles early. The probing algorithm's jerk is slightly higher compared to the others due to its probing actions. Both CC-MPC and the no-probing variant show larger gaps to vehicle 2 and smaller gaps to the trailing vehicle, indicating consistent merging behind all vehicles. In contrast, our active probing algorithm maintains moderate gaps to both vehicles 1 and 2, demonstrating its ability to identify and exploit merging opportunities between vehicles. 
% \jd{can we actually verify how many times the ego merged behind the last car? Does probing make ego vehicles more decisive compared to passive methods - explain? this maybe an interesting addition to the table as consistently merging last is easier because of low risk but will not always be realistic in a continuous traffic situation} \jd{do you have the distribution of the merging completion times?}

\begin{table}
  \centering
  \caption{Monte Carlo results for lane change scenario}\label{tab:montecarlo-merging}
  \begin{tabular}{|| c | c | c | c ||}
    \hline
    \textbf{Feature}  & \textbf{CC-MPC} & \textbf{No probing} & \textbf{Probing} \\ \hline
    Success rate & 62\% & 82\% & 98\% \\ \hline
    Time to merge($s$) & 9.271 & 7.215 & 6.871 \\ \hline
    Gap to vehicle 2($m$)  & 5.469 & 4.685 & 3.019 \\ \hline
    Gap to vehicle 1($m$) & 1.473 & 2.031 & 2.836 \\ \hline
    Velocity($m~s^{-1}$) & 4.378 & 5.976 & 5.829 \\ \hline
    Longitudinal jerk($m~s^{-3}$) & -0.201 & -0.191 & -0.208 \\ \hline
    Angular jerk($\text{rad}~s^{-3}$) & 2 $\times 10^{-3}$ & 1 $\times 10^{-2}$ & 1 $\times 10^{-2}$ \\  \hline
\end{tabular}
\end{table}

\subsection{Monte Carlo simulations - intersection scenario}
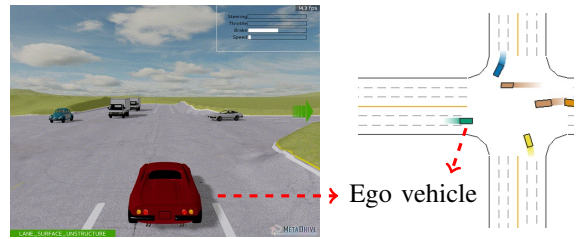
\begin{figure}[tbh]
  % \hspace*{1cm}
  \begin{tikzpicture}
    \node (img1) at (-1,0)  {\includegraphics[width=0.23\textwidth]{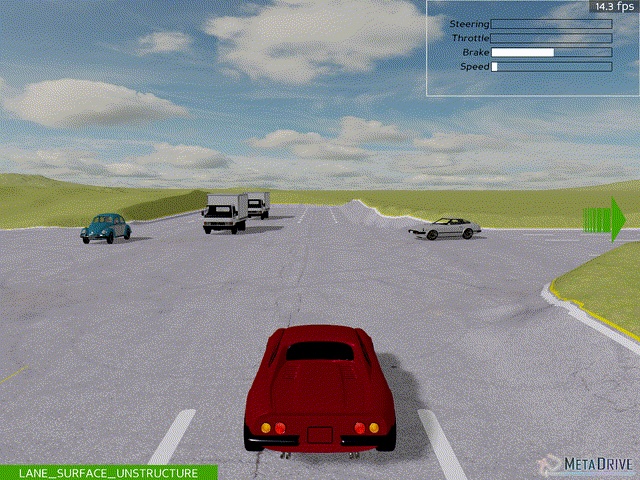}};
    \node (img2) at (3,0)  {\includegraphics[width=0.16\textwidth]{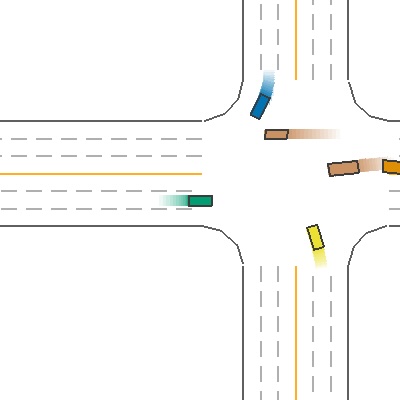}};
    \draw[->, dashed,very thick, red] (-0.30,-1) -- (1.3,-1);
    \draw[->, dashed, very thick, red] (3,-0.1) -- (2.8,-0.7);
      \node at (2.3,-1.0) {Ego vehicle};
  \end{tikzpicture}
  % \vspace*{-1em}
  \caption{Unsignalized intersection scene with $N=7$ vehicles. Ego vehicle's perspective (left), and its corresponding bird's eye view (right).}
  \vspace*{-0.3cm}
  \label{fig:intersection-scene}
  \end{figure}

We evaluate our algorithm in an unsignalized intersection scenario with $N=7$ vehicles controlled by MetaDrive's built-in reinforcement learning controllers. Note that the ego vehicle still maintains the same model of the other vehicle's behavior~\eqref{eqn:ego-perspective-agent-reward-model}, whereas the true behavior is prescribed by MetaDrive's in-built controller. We conduct Monte Carlo simulations comparing our approach against CC-MPC and the no-probing variant, defining success as safely exiting the intersection. Figure~\ref{fig:intersection-scene} illustrates one episode of the intersection scenario with the ego vehicle navigating through the intersection from two perspectives. 
% \jd{i am curious about the failure cases. presenting a failure case analysis or some explanation about it can be interesting for the readers}

Table~\ref{tab:montecarlo-intersection} summarizes the results. Our active probing algorithm achieves both the highest success rate and collision-free operation while enabling faster intersection crossing. Despite maintaining smaller gaps to other vehicles due to its probing behavior, the algorithm demonstrates safe operation with higher average velocities and jerk values comparable to the CC-MPC and the no probing algorithm. 
% \jd{do you have the acceleration/control values that acn be added to theb table? generally probing versions have higher control input i.e. higher suvccess and faster execution comes at the cost of expending higher control input due to probing maneuvers}.

% \jd{we talk about computational efficieny in the earlier sections but we havent complemented that with any numbers in the results. do you have higher precision values for angular jerk?  can we add some information about that? also talking about the scaling as prediction modes increase might be interesting to add if space is avaialble}

\begin{table}
  \centering
  \caption{Monte Carlo results for intersection scenario}\label{tab:montecarlo-intersection}
  \begin{tabular}{|| c | c | c | c ||}
    \hline
    \textbf{Feature}  & \textbf{CC-MPC} & \textbf{No probing} & \textbf{Probing} \\ \hline
    Success rate & 89\% & 93\% & 96\% \\ \hline
    Collision rate & 3\% & 0.5\% & 0.0\% \\ \hline
    Time to cross($s$) & 12.21 & 11.96 & 11.71 \\ \hline
    Gap to other vehicle($m$)  & 6.71 & 5.93 & 6.12 \\ \hline
    Velocity($m~s^{-1}$) & 5.049 & 5.3624 & 5.421 \\ \hline
    Longitudinal jerk($m~s^{-3}$) & -0.174 & -0.156 & -0.163 \\ \hline
    Angular jerk($\text{rad}~s^{-3}$) & 1 $\times 10^{-2}$ & 1 $\times 10^{-2}$ & 1 $\times 10^{-2}$ \\  \hline
\end{tabular}
\end{table}

% \subsection*{Remarks on computational efficiency}
% We solve all the nonlinear MPC problems in
% casadi \cite{JAEA-etal:2019_casadi}. 
% \begin{itemize}
% \item Computation time
% \item Speed planning vs full trajectory planning, filtering unlikely trajectories 
% \item Dependence on number of futures, vehicles, etc.
% \end{itemize}
\section{Conclusion}
We present a trajectory planning algorithm that combines multimodal predictions with active probing to estimate and adapt behaviors in complex scenarios. Our approach integrates safety-aware planning with information maximization objectives. This enables effective behavior estimation through strategic probing actions. Experiments in both lane change and intersection scenarios demonstrate that our algorithm outperforms state-of-the-art planners in terms of success rate and safety. Future work will focus on more general behavior models and computational optimizations through prediction filtering and simplified speed planning. 
%%%%%%%%%%%%%%%%%%%%%%%%%%%%%%%%%%%%%%%%%%%%%%%%%%%%%%%%%%%%%%%%%%%%%%%%%%%%%%%%

\bibliographystyle{unsrt}
\bibliography{refs}

%%%%%%%%%%%%%%%%%%%%%%%%%%%%%%%%%%%%%%%%%%%%%%%%%%%%%%%%%%%%%%%%%%%%%%%%%%%%%%%%

%%%%%%%%%%%%%%%%%%%%%%%%%%%%%%%%%%%%%%%%%%%%%%%%%%%%%%%%%%%%%%%%%%%%%%%%%%%%%%%%
% \section{APPENDIX}
% \subsection{Optimal transport and Wasserstein}\label{appn:wasserstein}
% Preliminaries on 2-Wasserstein distance and KL-divergence for Gaussians.

% %%%%%%%%%%%%%%%%%%%%%%%%%%%%%%%%%%%%%%%%%%%%%%%%%%%%%%%%%%%%%%%%%%%%%%%%%%%%%%%%

% \section*{Acknowledgements}

% We thank people for discussions.

\end{document}